\documentclass[twoside]{article}

\usepackage[preprint]{aistats2027}
\usepackage{amsmath,amssymb,amsfonts}
\usepackage{amsthm}

\usepackage{graphicx}
\usepackage{booktabs}
\usepackage{array}

\usepackage{wrapfig}

\usepackage{multirow}
\usepackage{wrapfig}

\usepackage{url}
\usepackage{csquotes, graphicx, xcolor, float}
\usepackage{bbm}
\usepackage{xcolor}
\usepackage{mathtools}
\usepackage{tikz}

\usepackage{natbib}
\usepackage[colorlinks=true, allcolors=blue]{hyperref}

\usepackage[utf8]{inputenc} 
\usepackage[T1]{fontenc}    
\usepackage{hyperref}       
\usepackage{url}            
\usepackage{booktabs}       
\usepackage{amsfonts}       
\usepackage{nicefrac}       
\usepackage{microtype}      
\usepackage{xcolor}         

\usepackage[capitalize,nameinlink]{cleveref}
\crefname{section}{Sec.}{Sections}
\crefname{thm}{Thm.}{Theorems}
\crefname{appendix}{App.}{Appendices}
\crefname{algorithm}{Alg.}{Algorithms}
\crefname{equation}{Eq.}{Eqs.}
\crefname{figure}{Fig.}{Figs.}
\crefname{prop}{Prop.}{Props.}
\creflabelformat{equation}{#2\textup{#1}#3} 

\definecolor{TolMutedBlue}{HTML}{332288}
\definecolor{TolMutedGreen}{HTML}{117733}
\definecolor{TolMutedPurple}{HTML}{AA4499}
\definecolor{TolMutedRed}{HTML}{BB3322}
\hypersetup{
    colorlinks=true,
    citecolor=TolMutedBlue,%
    linkcolor=TolMutedGreen,%
    urlcolor=TolMutedPurple,%
}

\newcommand{\dkl}[2]{\mathbb{D}_{\text{KL}}(#1 \, \|\, #2)}

\newcommand\cut[1]{}

\newcommand{\squishlist}{
   \begin{list}{$\bullet$}
    { \setlength{\itemsep}{0pt}      \setlength{\parsep}{3pt}
      \setlength{\topsep}{3pt}       \setlength{\partopsep}{0pt}
      \setlength{\leftmargin}{1.5em} \setlength{\labelwidth}{1em}
      \setlength{\labelsep}{0.5em} } }

\newcommand{\squishlisttwo}{
   \begin{list}{$\bullet$}
    { \setlength{\itemsep}{0pt}    \setlength{\parsep}{0pt}
      \setlength{\topsep}{0pt}     \setlength{\partopsep}{0pt}
      \setlength{\leftmargin}{2em} \setlength{\labelwidth}{1.5em}
      \setlength{\labelsep}{0.5em} } }

\newcommand{\squishend}{
    \end{list}  }

{}
\newtheorem{prop}{Proposition}{}
{}
\newcommand{\real}{\mbox{$\mathbb{R}$}}

\newcommand{\gauss}{\mbox{${\cal N}$}}

\newcommand{\myvec}[1]{\mbox{$\mathbf{#1}$}}
\newcommand{\myvecsym}[1]{\mbox{$\boldsymbol{#1}$}}

\newcommand{\vzero}{\mbox{$\myvecsym{0}$}}

\newcommand{\vmu}{\mbox{$\myvecsym{\mu}$}}

\newcommand{\vlambda}{\mbox{$\myvecsym{\lambda}$}}

\newcommand{\vphi}{\mbox{$\myvecsym{\phi}$}}

\newcommand{\vpsi}{\myvecsym{\psi}}

\newcommand{\vtheta}{\mbox{$\myvecsym{\theta}$}}

\newcommand{\vsigma}{\mbox{$\myvecsym{\sigma}$}}
\newcommand{\vSigma}{\mbox{$\myvecsym{\Sigma}$}}

\newcommand{\ve}{\mbox{$\myvec{e}$}}

\newcommand{\vm}{\mbox{$\myvec{m}$}}

\newcommand{\vt}{\mbox{$\myvec{t}$}}
\newcommand{\vu}{\mbox{$\myvec{u}$}}

\newcommand{\vx}{\mbox{$\myvec{x}$}}

\newcommand{\vy}{\mbox{$\myvec{y}$}}

\newcommand{\vz}{\mbox{$\myvec{z}$}}

\newcommand{\vI}{\mbox{$\myvec{I}$}}

\newcommand{\vS}{\mbox{$\myvec{S}$}}
\newcommand{\vT}{\mbox{$\myvec{T}$}}

\newcommand{\vX}{\mbox{$\myvec{X}$}}

\newcommand{\Var}{\mbox{$\operatorname{Var}$}}

\begin{document}

%

%
\runningauthor{Sklaviadis, M\"{o}llenhoff, Khan, Martins, Figueiredo}

\twocolumn[

\aistatstitle{The Fenchel-Young Generalization of Bayesian Principles}

\aistatsauthor{%
  Sophia Sklaviadis\textsuperscript{1} \\
  {\normalfont\texttt{ssklaviadis@gmail.com}}
    \And
  Thomas M\"{o}llenhoff\textsuperscript{2}
  \And
  Mohammad Emtiyaz Khan\textsuperscript{2,3}
  \AND
  Andr\'{e} F. T. Martins\textsuperscript{1}
  \And
  M\'{a}rio A. T. Figueiredo\textsuperscript{1}

}

\aistatsaddress{%
  \textsuperscript{1}Instituto de Telecomunica\c{c}\~{o}es and \\
  Instituto Superior T\'{e}cnico \\
  Universidade de Lisboa, Portugal
  \And
  \textsuperscript{2}RIKEN Center for \\
  Advanced Intelligence Project \\
  Tokyo, Japan
  \And
  \textsuperscript{3}TU Darmstadt and hessian.AI \\
  Darmstadt, Germany
}
]

\begin{abstract}
Conjugate Bayes plays a key role in machine learning, ensuring computationally simple posterior updates. We generalize the variational principle behind this computation by using the Fenchel-Young (FY) inequality, which lets us define a generalized evidence lower bound beyond the KL geometry, and at equality, characterizes new generalized posteriors. A generalized notion of conjugacy follows, defined as closure of prior-potential families that are affine under translation by the loss. We show that such families not only preserve additive canonical updating, but also posterior inference with parameter-dependent supports and tuneably heavy tails. This combination is not feasible with standard conjugate Bayes. We derive a new FY generalization of expectation maximization (EM), which recovers classical and Hard EM, and amortized versions. We show examples, including generalized Bayesian linear regression with closed-form bounded support and heavy-tailed posteriors, and a Gaussian mixture with sparse cluster assignments.
\end{abstract}

\begin{figure}
\centering
    \centering
    \includegraphics[width=\linewidth]{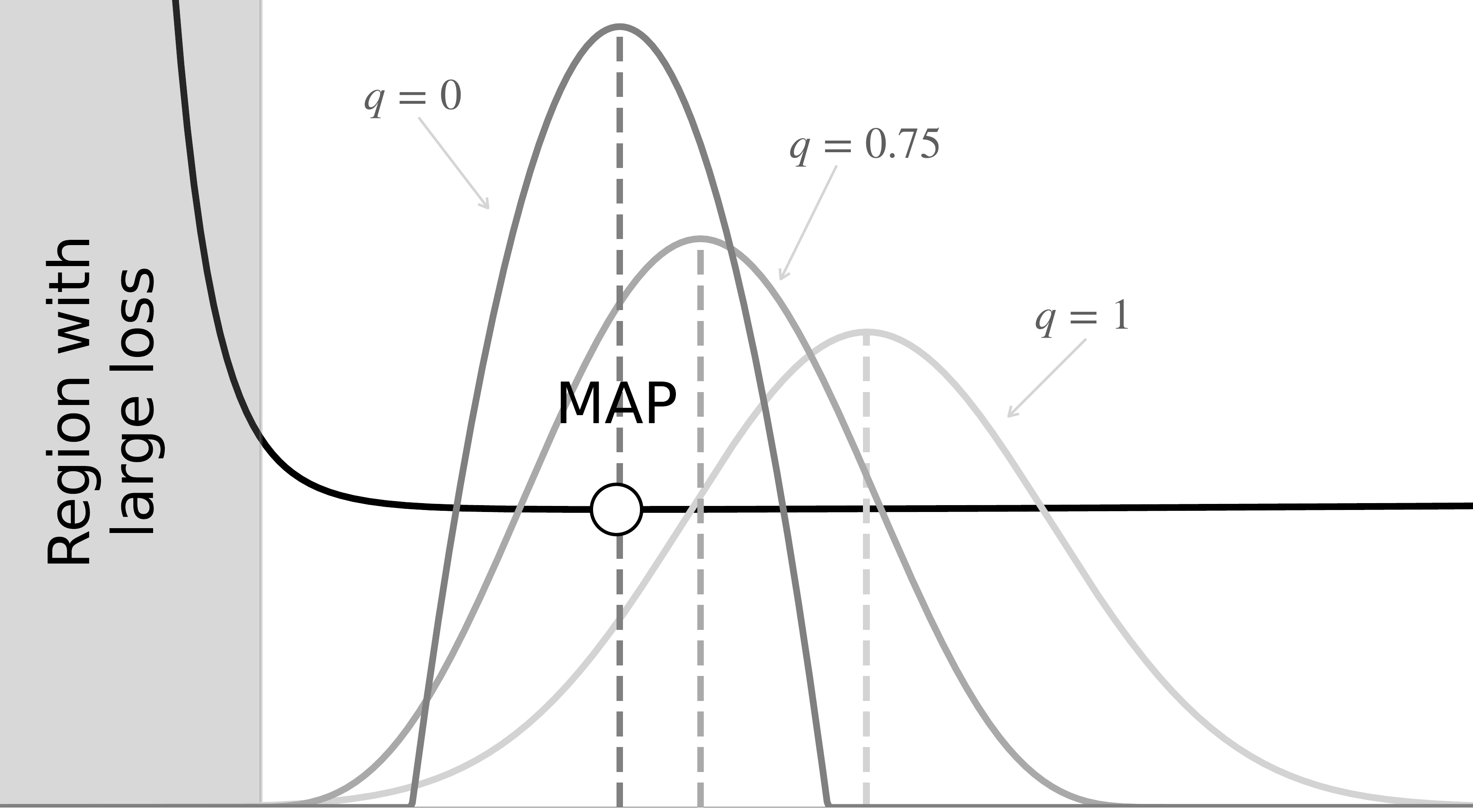}
    \caption{By changing the Kullback-Leibler divergence underlying Bayesian principles to a Fenchel-Young discrepancy, we retain Bayes' conceptual elegance but get generalizations with interesting properties. FY~discrepancies induced by Tsallis negentropies lead to $q$-Gaussian posteriors, which can be closer or even at the MAP estimate without collapsing to a Dirac measure.
    }
    \label{fig:sparselogistic}
\end{figure}

\section{Introduction}
\label{sec:intro}

Given a prior $\pi_0(\vtheta)$, about some unknown $\vtheta$, a likelihood relates it to observations, and Bayes' rule defines a posterior distribution over the unknown through multiplication and normalization. Equivalently, Bayesian inference is an unrestricted optimization over the space of distributions; Gibbs' variational principle already admits a general loss $\ell(\vtheta)$:
\begin{align}
\label{eq:gibbs_variational_principle}
    \pi^\star = \arg\min_{\pi\in\mathcal P} \underset{\text{\vtheta}\sim\pi}{\mathbb E} [\ell(\vtheta)] + \dkl{\pi}{\pi_0}.
\end{align}
The term $\dkl{\pi}{\pi_0} = \mathbb E_\pi \log (\pi/\pi_0) $ is the Kullback-Leibler (KL) divergence. When $\ell(\vtheta)$ is the negative log-likelihood, and $\mathcal P$ is the space of all densities or mass functions, $\pi^\star$ is the Bayes posterior \citep{williams1980bayesian}.
The variational form of posterior inference in \cref{eq:gibbs_variational_principle} appears in many classic and more recent works \citep{jaynes1957information,jaynes88, csiszar1975divergence, donsker1975asymptotic,Zellner1988,zhang1999theoretical,catoni2007pac, bissiri2016general, alquier2016properties, jewson2022general, knoblauch2022optimization, khan2023bayesian}.

\citet{knoblauch2022optimization} propose a broad generalization of variational Bayes, addressing misspecified models and computational constraints. 
Their Rule of Three (RoT) independently relaxes three pieces of the evidence lower bound (ELBO), i.e. of the negative argument in the right hand side of \cref{eq:gibbs_variational_principle}: 
One may vary the loss $\ell(\vtheta)$, replace the KL by any other divergence, and optimize over a subset $\mathcal Q \subseteq \mathcal P$. 
This approach differs from alternative generalizations of variational Bayes which restrict posterior candidates to a family $\mathcal Q$ while minimizing general divergences between candidate posteriors and the Bayes posterior \citep{minka2005divergence,hernandez2016black,li2016renyi}.

While the RoT relies on Gibbs' variational principle, its modular generalization of Bayesian inference can result in infinite objectives; e.g.
if the prior has bounded support and $\mathcal Q$ is a family of Gaussian densities with full support, then the KL is \(+\infty\) for every candidate in \(\mathcal Q\), and the objective is uninformative on that family. 
Such problems can be avoided by an alternative approach to generalizing Bayesian inference beyond the KL divergence.  
Our question is not whether the KL can be generalized, rather we ask how to preserve the Bayesian coupling between prior and posterior, as well as computationally efficient conjugate updating.

Our key idea stems from the fact that the KL term can be expressed more generally by using the Fenchel-Young (FY) inequality. In particular, the KL term in \cref{eq:gibbs_variational_principle} is the gap between the two sides of the FY inequality $\Omega(\pi)+\Omega^*(\log\pi_0) \geq \mathbb E_\pi[\log\pi_0]$, where $\Omega$ is negative Shannon entropy and $\Omega^*$ is its convex conjugate, the log-normalizing function.
Then, applying the Fenchel-Young equality condition at $\log\pi_0-\ell$ identifies the Gibbs posterior $\pi^\star\propto\pi_0\exp(-\ell)$ as the solution to \cref{eq:gibbs_variational_principle}.
The structure that generates the KL also induces generalized Fenchel-Young posterior families, and motivates a novel characterization of conjugate families, casting classical results of \citet{diaconis1979conjugate, Barndorff-Nielsen} in a broader context.

We introduce the Fenchel-Young (FY) generalization of variational Bayes which leads to new inference schemes with interesting properties (see~\Cref{fig:sparselogistic}), but preserves the Bayesian coupling of prior representation, discrepancy, and induced posteriors. 
We use a convex function $\Omega$ to define a Fenchel-Young geometry \citep{amari2016information,Blondel2020, nielsen2022statistical}, associated with a discrepancy; KL and Bayes are recovered when the convex function is the negative Shannon entropy.
We represent unnormalized prior information by a potential $\eta_0$, paired with a distribution through its expectation.
When $\eta_0\in\partial\Omega(\pi_0)$, the FY discrepancy is exactly the subgradient Bregman discrepancy from $\pi_0$.
The loss, $\ell$, translates the unnormalized prior potential to $\eta_0-\ell$, and the same convex $\Omega$ that defines the discrepancy between variational posterior and prior potential, induces the generalized posterior as an unrestricted optimizer. 
The FY gap at the translated potential is exactly the gap between the FY evidence and its variational lower bound; equality characterizes the generalized posterior.

We characterize conjugate updating through closure of affine potential families under these translations.
While different choices of $\Omega$ lead to generalized (conjugate) posteriors, updating always amounts to addition in the dual space. 
A finite convex conjugate value $\Omega^*$ at $\eta_0$ does not presuppose an induced prior density, as the constant-potential example below illustrates.
We derive FY expectation maximization, which alternates between local posterior updates, and fitting the observation model and prior potential. 
We illustrate FY Bayes using Tsallis entropies in bounded support and heavy tailed conjugate linear regression, FY expectation maximization for Gaussian mixture clustering with sparse posterior assignments, and VAEs with FY posteriors.

\section{Fenchel-Young Bayes}

We review classical variational Bayes and conjugate families in \cref{appendix:vcb}. 
Throughout, we will use a scoring function or ``prior potential'' to generalize the log prior. 
We first define Fenchel-Young (FY) discrepancies, then show that the data update the prior potential by subtracting the loss.
This is a systematic generalization of Bayesian inference, through which we characterize FY posteriors as unrestricted optimizers of the generalized variational objective, and extend conjugacy beyond parametric exponential families \citep{diaconis1979conjugate, Barndorff-Nielsen}.

\subsection{The FY Generalization of the KL}

The KL divergence is the Fenchel-Young gap generated by Shannon negentropy.
We use the same convex dual construction with general convex generator functions, pairing a probability distribution with a scoring function. 
Fenchel-Young discrepancies have been used as losses in supervised learning \citep{amari2010information, Blondel2020,martins2022sparse}, and are closely related to proper scoring rules \citep{dawid2007geometry}. 
Here, we use this structure to characterize unrestricted posterior updating and generalized conjugacy.

We work on a measurable space $\Theta$ with a fixed reference measure $h_\Theta$. We consider measurable potentials and losses, and let $\mathcal P$ be the set of probability densities under which these functions are integrable. We get mass functions using the counting measure, and our continuous examples use the Lebesgue measure. 
We pair a density $\pi$ with a scoring function $\eta$ through $\langle\pi,\eta\rangle=\mathbb E_\pi[\eta]$.
Subgradients are understood through supporting inequalities under this pairing.

Let $\Omega:\mathcal P\to\mathbb R\cup\{+\infty\}$ be proper and convex, with Fenchel conjugate $\Omega^*$ defined by the supremum in \cref{eq:Omega_star}, and let  
$\eta:\Theta\to\mathbb R$. 
The FY discrepancy generated by $\Omega$ quantifies the incompatibility of potential $\eta$, and density or mass function $\pi$, in terms of the FY gap:
\begin{align}\label{eq:LOmega}
\mathbb D_\Omega(\eta;\pi)
&= \Omega^*(\eta) + \Omega(\pi) - \langle\pi,\eta\rangle \\
&= \Omega^*(\eta) + \Omega(\pi) - \underset{\text{\vtheta}\sim\pi}{\mathbb E} [\eta(\vtheta)],  \nonumber
\end{align}
for the pairing \( \langle\pi,\eta\rangle = \mathbb E_{\text{\vtheta}\sim\pi} [\eta(\vtheta)]\).
The Fenchel-Young inequality ensures non-negativity,
\begin{align} \label{eq:Omega_star} 
\Omega^*(\eta) = \sup_{\pi\in\mathcal P} \langle\pi,\eta\rangle-\Omega(\pi)  \quad  
\Longrightarrow \quad
\mathbb D_\Omega(\eta;\pi) \geq 0.
\end{align}
Moreover, $\mathbb D_\Omega(\eta;\pi)=0$ implies that
\[ 
        \pi \in \underset{\pi\in\mathcal P}{\arg\max} \,\,\,
        \langle\pi,\eta\rangle-\Omega(\pi)
        \quad\implies\quad
        \pi \in \partial\Omega^*(\eta).
\]
That is, $\pi$ is the subgradient of the Fenchel conjugate evaluated at $\eta$.
Finally, because all $\pi\in\mathcal P$ integrate to one, for any constant $C$
\begin{equation}
\mathbb D_\Omega(\eta+C;\pi) = \mathbb D_\Omega(\eta;\pi).
\label{eq:fy_shift_invariance}
\end{equation}
When $\Omega$ is the negative of the Shannon entropy $\mathcal H(\pi) = -\int_\Theta \pi \log \pi$, then the FY gap reduces to the KL: 
\[ 
\mathbb D_{-\mathcal H}(\eta;\pi)
= \mathbb D_{\rm KL} \left( \pi ~\Bigg\|~ \frac{\exp(\eta)} {\int_\Theta\exp(\eta(\vtheta))} \right).
\]


\subsection{FY Evidence Lower Bound} 

For observed data $\vy$, loss $\ell(\vy;\vtheta)$, and prior potential $\eta_0\in\text{dom}\Omega^*$, we define the FY variational free energy
\begin{align}\label{eq:fy_free_energybeta} 
F_{\ell,\Omega,\beta}(\pi,\eta_0;\vy) = \underset{\text{\vtheta}\sim\pi}{\mathbb E} \left[ \ell(\vy;\vtheta) \right] + \beta\mathbb D_\Omega(\eta_0;\pi). 
\end{align} 
The potential $\eta_0$ need not induce a prior density. The parameter $\beta>0$ controls the relative weight of the expected loss and the FY discrepancy. In the Shannon case, $\beta$ is the temperature of the
Gibbs posterior \citep{bissiri2016general}; it appears in $\beta$-VAEs \citep{Higgins2017betaVAELB}, and can be interpreted as a rate-distortion trade-off \citep{alemi2018information}. When we omit $\beta$, we assume $\beta=1$.

With $\beta=1$, and using \cref{eq:LOmega}, we can write \cref{eq:fy_free_energybeta} as:
\begin{align} 
\label{eq:genconjbayes}
F_{\ell,\Omega}(\pi,\eta_0;\vy) 
&= \mathbb D_\Omega \left( \eta_0-\ell(\vy;\cdot); \pi \right)  \\ 
& \ \ + \Omega^*(\eta_0) - \Omega^* \left( \eta_0-\ell(\vy;\cdot) \right). \nonumber
\end{align} 
The last two terms do not depend on $\pi$. So, when the infimum over $\mathcal P$ is attained, the FY posterior satisfies 
\begin{align}  
\label{eq:opt_post} 
\pi^\star \in \partial\Omega^* \left( \eta_0-\ell(\vy;\cdot) \right).
\end{align} 
This shows that the generalized FY posterior is characterized by the subgradient of $\partial\Omega^*$ evaluated at the translated score, that is, the affine operation where $\eta_0$ is shifted by subtracting the loss.
For parametric conjugacy, it is then sufficient that $\eta_0-\ell$ is exactly represented in the chosen affine family.

The unrestricted infimum defines the negative Fenchel-Young evidence, 
\begin{align} 
J_{\ell,\Omega}(\vy,\eta_0) 
= \inf_{\pi\in\mathcal P} F_{\ell,\Omega}(\pi,\eta_0;\vy), 
\end{align} 
and the variational gap is exactly
\[
F_{\ell,\Omega}(\pi,\eta_0;\vy)-J_{\ell,\Omega}(\vy,\eta_0)
=\mathbb D_\Omega\bigl(\eta_0-\ell(\vy;\cdot);\pi\bigr)\geq0.
\]
Thus, $-F_{\ell,\Omega}$ is an FY evidence lower bound (FYELBO). If the unrestricted FY posterior belongs to $\mathcal Q \subseteq\mathcal P$, the bound is tight; otherwise, optimization over $\mathcal Q$ amounts to a variational approximation. 
\Cref{eq:genconjbayes} shows that replacing the FY posterior by an approximation $\pi$ increases the variational free energy by $\beta\mathbb D_\Omega(\eta_0-\ell(\vy;\cdot)/\beta;\pi)$. 
For Shannon negentropy, this is $\beta$ times the KL divergence to the Gibbs posterior, as in the PAC-Bayes analysis of \citet{alquier2016properties}.

The standard Bayes' update can be obtained as a special case: We set $\Omega=-\mathcal H$ and $\ell(\vy;\vtheta) = -\log p(\vy\mid\vtheta)$. Additionally, we define the scoring function to be the log-prior, that is, we define $\eta_0(\vtheta) = \log\pi_0(\vtheta)$. Thus, the FY evidence and posterior reduce, respectively, to the log marginal likelihood and the Bayes posterior
\[ 
-J_{\ell,\Omega}(\vy,\eta_0) = \log p(\vy), \qquad \pi^\star = p(\vtheta\mid\vy). 
\] 
In classical Bayesian analysis, the evidence is used for model comparison and hypothesis testing \citep{Bayes_factors}. 
In general, the FY evidence is not a marginal likelihood and does not assume equivalent interpretation. 
Nevertheless, $-J_{\ell,\Omega}$ is a generalized comparison criterion, and evidence-based criteria have been useful in deep learning even when the model is not treated as fully Bayesian \citep{immer2021scalable}.

\subsection{FY Conjugate Bayes}

To obtain a finite dimensional family, we take \smash{$\vT:\Theta\to\mathbb R^P$}, $\vlambda\in\mathbb R^P$, and restrict the score to
\begin{align}
\label{eq:lin_score}
\eta_{\text{\vlambda}}(\vtheta) = \left\langle \vlambda,\vT(\vtheta) \right\rangle. 
\end{align}
Let $\Lambda$ be the set of $\vlambda$ for which $\Omega^*(\eta_{\text{\vlambda}})$ is finite, and the supremum defining $\Omega^*$ is attained.
Whenever the supremum is attained 
at \(\pi_{\text{\vlambda}}\), that density minimizes \(\Omega\) among all densities with the same moments of \(\vT\) (see \cref{app:maxent}). 
Using the Shannon negative entropy as \(\Omega\) we can derive the exponential family (see \cref{appendix:vcb}). 
Thus, $\Omega$ and $\vT(\vtheta)$ induce a generalized maximum entropy family, $\pi_{\text{\vlambda}} \in \partial \Omega^*(\eta_{\text{\vlambda}})$, with canonical domain $\Lambda$.

This maximum entropy characterization identifies the parametric families induced by the FY Bayes.
The moment constraints are not additional empirical constraints imposed
on the posterior update, rather they characterize the families in question.

For a dataset $\mathcal D$ with loss $\ell_{\mathcal D}(\vtheta)=\ell(\mathcal D;\vtheta)$, and
a prior potential $\eta_{\text{\vlambda}_0}$, consider losses satisfying
\begin{equation}
-\ell_{\mathcal D}(\vtheta)
= \left\langle \vt(\mathcal D), \vT(\vtheta) \right\rangle - c(\mathcal D),
\label{eq:fy_affine_loss}
\end{equation}
where $\vt(\mathcal D)\in\mathbb R^P$ is a data-dependent vector, $\vT(\vtheta)$ defines the basis of the space of potentials, and $c(\mathcal D)$ does not depend on $\vtheta$. 
The translated potential is
\begin{align}
\eta_{\text{\vlambda}_0}(\vtheta) - \ell_{\mathcal D}(\vtheta)
&= \eta_{\text{\vlambda}_{\mathcal D}}(\vtheta) - c(\mathcal D),
\end{align}
where, by \cref{eq:fy_shift_invariance}, the additive constant does not affect the FY posterior. 
Thus, whenever $\vlambda_{\mathcal D}\in\Lambda$, the density $\pi_{\text{\vlambda}_{\mathcal D}}$ is an exact FY posterior, with $\vlambda_{\mathcal D}=\vlambda_0+\vt(\mathcal D)$, as in classical conjugate Bayes (\cref{appendix:vcb}).

The family $\{\pi_{\text{\vlambda}}:\vlambda\in\Lambda\}$ is conjugate to the class of losses with the affine form in \cref{eq:fy_affine_loss}, and $\vlambda+\vt(\mathcal D)\in\Lambda$ for every $\vlambda\in\Lambda$ and every dataset $\mathcal D$ in the class of losses.
If the loss decomposes over observations, then \( \vt(\mathcal D) = \sum_{i=1}^N \vt(\mathcal D_i) \).
Thus, different choices of $\Omega$ change the density associated with an affine score, but not the additive coefficient update. 
For the weighted objective with $\beta\neq1$, the update is $\vlambda_{\mathcal D} = \vlambda_0+\vt(\mathcal D)/\beta$.
We derive the family from the same $\Omega$ that defines the discrepancy, rather than choosing it independently as a variational restriction. 
For affine prior potentials and losses, fixing the moments of $\vT$ fixes every objective term except $\Omega(\pi)$. 
Minimizing $\Omega$ at fixed moments therefore minimizes the full objective among densities with those moments. 
Under the conjugacy conditions above, this family contains the unrestricted posterior.

\subsection{Tsallis Entropies} 
Setting $\Omega$ to be the Tsallis $(2-q)$-negentropy, lets us derive interesting posteriors with bounded and sparse supports, as well as heavy tails
\begin{align}\label{eq:tsallis}
\Omega_{2-q}(\pi)
=
\begin{cases}
\displaystyle
\frac{\underset{\text{\vtheta}\sim\pi}{\mathbb E} \left[\pi(\vtheta)^{1-q}-1\right]}{(2-q)(1-q)},
& q\neq 1,\\[2mm]
\displaystyle
\underset{\text{\vtheta}\sim\pi}{\mathbb E} [\log\pi(\vtheta)],
& q=1.
\end{cases}
\end{align}
Deformed $q$-exponential families attain $\Omega^*_{2-q}(\eta_{\text{\vlambda}})$ when they have finite moments of $\vT$ and finite $\Omega_{2-q}$:
\begin{equation}
\pi_{\text{\vlambda}}(\vtheta)
= \exp_q\left( \langle\vlambda,\vT(\vtheta)\rangle - A_{2-q}(\vlambda) \right), 
\end{equation}
where $\exp_q (u) =  [1 + (1-q) u]_+^{1/(1-q)}$ and $A_{2-q}$ is the normalizing constant or deformed cumulant. 
This is the Tsallis instance of the maximum entropy characterization above; the proof is in \cref{ssec:tsallis_maxent}.

In general, the deformed cumulant is determined by
\[
\int_\Theta
\exp_q \left(\langle\vlambda,\vT(\vtheta)\rangle - A_{2-q}(\vlambda)\right) dh_\Theta(\vtheta)=1.
\]
For finite $\Theta = \{1,\ldots,K\}$, $A_{2-q}$ is determined by
\[
\sum_{k=1}^K \exp_q \left( \eta_k-A_{2-q}(\eta) \right) = 1,
\]
and the induced posterior is obtained through the $(2-q)$-entmax($\eta$) map defined by \citet{peters2019sparse}, which recovers softmax at $q=1$ and sparsemax at $q=0$ \citep{Martins2016}.

Ordinary exponential families have parameter-independent support \citep{Barndorff-Nielsen}. 
For $q<1$, the $q$-exponential families induced by Tsallis $(2-q)$-negentropies can instead have parameter-dependent bounded support. 
An ordinary exponential family may still provide a variational approximation when the FY objective is finite, but cannot represent the resulting posterior family's changing support. 

\paragraph{Finite regularizers for improper prior potentials.}
Let \smash{$\Theta=\mathbb R^D$} and \smash{$\eta_0=0$}.
For the Shannon negentropy, \smash{$\Omega^*_{-\mathcal H}(0)=+\infty$}.
For Tsallis negentropies with \smash{$q<1$},
\smash{$\Omega^*_{2-q}(0)=1/[(2-q)(1-q)]<\infty$}.
This value is approached by uniform densities on sets of increasing
volume, but is not attained by any density. 
Thus $\eta_0$ is not normalizable, but
\[
\mathbb D_{\Omega_{2-q}}(0;\pi) = \frac{\int_{\mathbb R^D}\pi(\vtheta)^{2-q} dh_\Theta(\vtheta)} {(2-q)(1-q)}
\]
is finite for $\pi\in\mathcal P\cap L^{2-q}(h_\Theta)$.
With Gaussian loss 
$\ell(\vy;\vtheta)=\|\vtheta-\vy\|_2^2/(2\sigma^2)$, the posterior is
\[
\pi^\star(\vtheta) = \exp_q \left( -\frac{\|\vtheta-\vy\|_2^2}{2\sigma^2}-A_{2-q} \right).
\]

\section{Fenchel-Young Expectation Maximization}\label{sec:FYEM}

Here, we show how FY Bayes leads to a natural generalization of the expectation maximization algorithm \citep{dempster1977maximum}. 
Let $\mathcal D=(y_1,\ldots,y_N)$ be the observations, and let $\theta=(\theta_1,\ldots,\theta_N)$ be associated local latent variables, so that $\theta_i$ is the unobserved random quantity associated with response $y_i$.
The latent space $\Theta$ may be discrete or continuous; $\pi_0$ and all $\pi_i$ are density or mass functions with respect to the same $h_\Theta$. 
The observation space need not be of the same type as the latent space. 
Conjugacy depends on the loss as a function of the latent variable. 

Let $\pi=(\pi_1,\ldots,\pi_N)\in\mathcal P^N$, where $\pi_i$ is a candidate posterior for $\theta_i$. 
The shared latent potential is $\eta_0$, and when it represents a probability distribution, we denote the density or mass function by $\pi_0$, so that \( \pi_0\in\partial\Omega^*(\eta_0)\). 
We add the single-observation free energies in \cref{eq:fy_free_energybeta}, one local posterior per observation, with  $(\vphi,\eta_0)$ shared, where $\vphi$ is the vector of observation model parameters.
In the FYEM criterion defined below we optimize these shared parameters as well:
\begin{align}
F_{\mathcal D}(\pi,\vphi,\eta_0)
&= \sum_{i=1}^N
\underset{\theta_i\sim\pi_i}{\mathbb E} \left[ \ell_{\text{\vphi}}(y_i;\theta_i) \right] + \mathbb D_\Omega(\eta_0;\pi_i),
\label{eq:joint_minimization}
\end{align}
where $\ell_{\text{\vphi}}(y_i;\theta_i)$ is the sample-wise loss.

Alternating minimization with respect to $\pi$ and $(\vphi,\eta_0)$ amounts to a novel EM-style algorithm. 
The E-step updates the local posteriors $\pi_i$, while the M-step updates the observation model parameters $\vphi$ and the shared prior potential $\eta_0$. 
We index iterates by $r$:
\paragraph{E-step:}
Given $(\vphi^{(r)},\eta_0^{(r)})$, for $i=1,\ldots,N$ update
\begin{align}
\pi_i^{(r+1)}
&\in \arg\min_{\pi_i\in\mathcal P}
\underset{\theta_i\sim\pi_i}{\mathbb E}
\left[ \ell_{\text{\vphi}^{(r)}}(y_i;\theta_i) \right] + \mathbb D_\Omega(\eta_0^{(r)};\pi_i)  \nonumber \\
&= \arg\min_{\pi_i\in\mathcal P}
\mathbb D_\Omega \left(  \eta_0^{(r)} - \ell_{\text{\vphi}^{(r)}}(y_i;\cdot); \pi_i \right).
\label{eq:e_step}
\end{align}
Whenever this minimum is attained,
\[
\pi_i^{(r+1)} \in \partial\Omega^*
\left( \eta_0^{(r)} - \ell_{\text{\vphi}^{(r)}}(y_i;\cdot) \right).
\]
With $\vphi^{(r)}$ fixed, if the negative loss has the affine form of  \cref{eq:fy_affine_loss}, and $\eta_0^{(r)}=\eta_{\text{\vlambda}_0^{(r)}}$, the E-step can be a conjugate update of each local latent distribution
\[
\pi_i^{(r+1)} = \pi_{\text{\vlambda}_0^{(r)}+\text{\vt}(y_i)},
\]
provided $\vlambda_0^{(r)}+\vt(y_i)\in\Lambda$.
Here $\vt(y_i)$ contains the coefficients of $-\ell_{\text{\vphi}^{(r)}}(y_i;\cdot)$ in $\vT$; these coefficients can change with $\vphi$.

\paragraph{M-step:}
Given $\pi^{(r+1)}$, update 
\begin{align}
&\left(\vphi^{(r+1)}, \eta_0^{(r+1)} \right) \in \\ 
&\arg\min_{\text{\vphi},\eta_0}
\sum_{i=1}^N  \underset{\theta_i\sim\pi_i^{(r+1)}}{\mathbb E}
\left[ \ell_{\text{\vphi}}(y_i;\theta_i) \right] + \mathbb D_\Omega
\left( \eta_0;\pi_i^{(r+1)} \right) . \nonumber
\label{eq:m_step}
\end{align}
When $\vphi$ and $\eta_0$ are optimized over a product parameter space, the minimization separates as
\begin{align}
\vphi^{(r+1)}
&\in \arg\min_{\text{\vphi}} \sum_{i=1}^N \underset{\theta_i\sim\pi_i^{(r+1)}}{\mathbb E} \left[ \ell_{\text{\vphi}}(y_i;\theta_i) \right],\\
\eta_0^{(r+1)}
&\in \arg\min_{\eta_0} \sum_{i=1}^N \mathbb D_\Omega \left( \eta_0;\pi_i^{(r+1)} \right) \nonumber \\
&= \arg\min_{\eta_0} \mathbb D_\Omega
\left(\eta_0; \bar\pi^{(r+1)} \right),
\end{align}
with
\[
\bar\pi^{(r+1)} = \frac1N \sum_{i=1}^N \pi_i^{(r+1)}.
\]
If $\partial\Omega(\bar\pi^{(r+1)})\neq\varnothing$, any $\eta_0^{(r+1)}\in\partial\Omega(\bar\pi^{(r+1)})$ solves the M-step, and we can set the shared prior to the mixture of local posteriors $\pi_0^{(r+1)}=\bar\pi^{(r+1)}$.
If the prior potential is restricted to an affine family with statistic $\vT$, any prior in the induced family satisfying $\mathbb E_{\pi_0^{(r+1)}}[\vT] = \mathbb E_{\bar\pi^{(r+1)}}[\vT]$ solves the M-step, matching moments, not necessarily the densities.
With Shannon negentropy and logarithmic loss, FYEM recovers EM.

Minimizing \cref{eq:joint_minimization} over local posteriors leaves an objective in $(\vphi,\eta_0)$, and keeping the posteriors fixed  gives an upper bound, because minimizing over them can only lower the value. 
The exact E-step makes this bound equal to the objective at $(\vphi^{(r)},\eta_0^{(r)})$. 
The M-step then minimizes the bound over $(\vphi,\eta_0)$. 
Together, these steps form a majorization-minimization (MM) iteration.

\subsection{Further Connections}

\paragraph{Proper scoring rules.} 
The proper scoring rule formulation of \cref{eq:joint_minimization} is an insightful source of inspiration for choosing $\Omega$'s, through the negative entropies defined by well-known proper scoring rules \citep{dawid2007geometry}; this connection is an exciting direction to explore in future work. 
Kernel scores, for example, can be used to define generators that depend on pairs of parameter values, rather than on each density value separately \citep[Sec.~6.3]{dawid2007geometry}.
Our posterior characterization and affine-potential conjugate updates apply to all these generators, while the FY formulation also permits prior potentials outside this representation.
Our contribution is to characterize the resulting posteriors as unrestricted optimizers and derive conjugate updates through affine translations of the potential.



\paragraph{Conjugate computation.} Suppose the loss does not satisfy \cref{eq:fy_affine_loss}, the moments of $\vT$ locally parametrize the induced family, and the expected loss is differentiable in these moments. 
A first-order approximation of the expected loss is affine in $\vT$. 
The resulting surrogate objective is solved by the conjugate update above, generalizing \citet{khan2017conjugate}. 
Only the expected loss is approximated. 
This first-order approximation does not always upper bound the FY posterior objective, so the exact E-step MM argument does not apply.

\paragraph{Amortized FY inference.} 
Amortization provides an alternative affine surrogate. 
An encoder network can learn to predict unknown canonical parameters, rather than computing them separately for each observation. 
The encoder is trained using the FYELBO evaluated at posteriors parametrized by $\vpsi$, \( \pi_i=\pi_{\text{\vpsi}}(\cdot\mid y_i) \).
Our VAE experiments optimize the encoder directly using reparametrized gradients and a fixed prior potential $\eta_0$. 
A score-function estimator may instead be used when a reparametrization is unavailable \citep{williams1992simple}.

\section{Examples}
\label{sec:experiments}

We illustrate FY Bayes: 
First, we derive generalized conjugate linear regression, with bounded support and heavy-tailed uncertainty. 
Then, we specialize FYEM to Gaussian mixture clustering; the Shannon negentropy recovers standard EM \citep{dempster1977maximum}. 
Restricting local posteriors to simplex vertices recovers classification or hard EM \citep{Celeux}, and the Tsallis negative entropy with $q<1$ characterizes a sparse E-step algorithm.\footnote{Our variant of EM is distinct from those proposed by \citet{dukkipati2016mixture} and \citet{inoue2013q}, which use dense mixtures of bounded support $q$-Gaussians.}
Finally, we fit small FY VAEs with bounded support posteriors on MNIST and FashionMNIST, using both sparse and standard Bernoulli observation models.\footnote{Tradeoffs between FY and KL regularization depend on the model, entropic index, and parametrization. 
In the reported experiments, additional computational cost of the FY regularizers relative to the corresponding KL baselines is negligible. Efficient specialized algorithms are available for sparsemax and $1.5$-entmax \citep{peters2019sparse}. 
We use reparametrized gradient estimators in all VAE experiments and balanced computational budgets for compared models.}

\subsection{Conjugate FY Linear Regression}

\begin{figure}[t!]
    \centering
    \includegraphics[width=\linewidth]{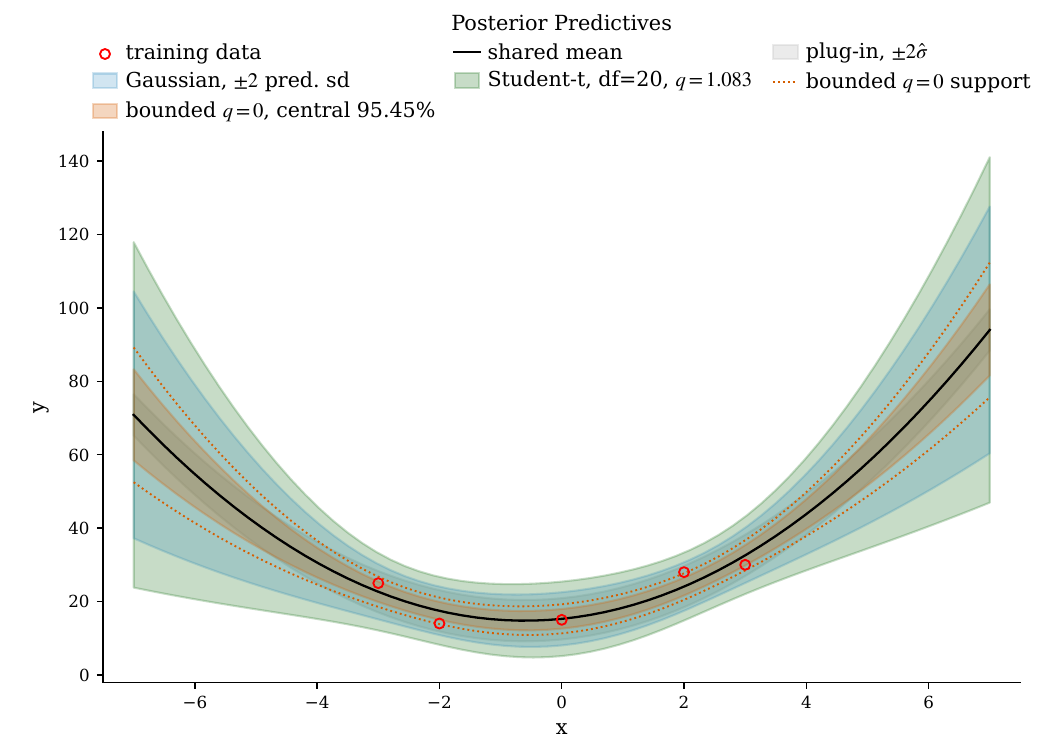}
        \caption{\small We extend \citet[Fig.~11.21]{murphy2022probabilistic}'s Bayesian linear regression, to illustrate closed-form generalized conjugate posteriors with bounded support (orange) and heavy Student-$t$ tails (green).  
        Shaded bands indicate the central $95.45\%$ intervals for future observations; the dotted curves correspond to the bounded predictive support.
        }
    \label{fig:predictives}
\end{figure}

We review ridge regression (c.f. \citet[Ch.~11]{murphy2022probabilistic}), and derive FY ridge regression by translating $q$-Gaussian prior potentials with the same quadratic loss, to exemplify computationally efficient, canonical parameter updates of generalized conjugate models. \cref{fig:predictives} illustrates bounded-support (orange) and heavy-tailed (green) predictive uncertainty.

Let $\mathcal D=(\vX,\vy)$, where $\vy\in\real^N$, $\vX\in\real^{N\times D}$, and $\vtheta\in\real^D$ be the unknown regression coefficient. Set
\[
\ell(\vtheta)
= \frac12\|\vy-\vX\vtheta\|_2^2,
\qquad
\pi_0(\vtheta)
= \gauss \left( \vtheta\mid\vzero,\delta^{-1}\vI \right),
\]
where $\delta>0$ is the prior precision.
Substituting \smash{$\pi_0(\vtheta)\propto \exp(-\delta\|\vtheta\|_2^2/2)$} into Bayes' rule 
\begin{align}
    \pi^\star(\vtheta)
    &\propto \pi_0(\vtheta) \exp \left(-\frac12 \|\vy - \vX\vtheta\|_2^2 \right) \\
    &\propto \exp \left( -\frac12 \|\vy-\vX\vtheta\|_2^2 - \frac{\delta}{2} \| \vtheta \|_2^2
     \right). 
\end{align}
Writing $\vS = \delta\vI+\vX^\top\vX$, 
\( 
\vm = \vS^{-1}\vX^\top\vy
\),
and completing the square, we get classical Gaussian posteriors, 
\(
\pi^\star(\vtheta) = \gauss \left( \vtheta\mid\vm,\vS^{-1} \right)
\) \citep[Ch.~11.7.2]{murphy2022probabilistic}.

Under the Tsallis $(2-q)$-negentropy from \cref{eq:tsallis}, quadratic scores induce the $q$-Gaussian generalized maximum entropy families (\cref{ssec:tsallis_maxent,app:reg}). We represent priors by quadratic potentials
\begin{align}
\eta_0(\vtheta)
&= -\frac{\delta}{2}\|\vtheta\|_2^2
= \left\langle
\vlambda_0,\vT(\vtheta)
\right\rangle,\\
\vT(\vtheta)
&=
\begin{pmatrix}
\vtheta\\
\vtheta\vtheta^\top
\end{pmatrix},
\qquad
\vlambda_0
=
\begin{pmatrix}
\vzero\\
-\frac{\delta}{2}\vI
\end{pmatrix}.
\end{align}
For the appropriate values of $q$, the supremum defining $\Omega_{2-q}^*(\eta_0)$ is attained and the induced prior is
\[
\pi_{\text{\vlambda}_0}(\vtheta)
= \exp_q
\left( \left\langle \vlambda_0,\vT(\vtheta) \right\rangle - A_{2-q}(\vlambda_0) \right),
\]
and $ A_{2-q}(\vlambda_0)$ is the deformed cumulant (\cref{app:reg}).

The negative quadratic loss is affine in $\vT$:
\begin{align}
-\ell(\vtheta)
&=
\left\langle
\begin{pmatrix}
\vX^\top\vy\\
-\frac12\vX^\top\vX
\end{pmatrix},
\vT(\vtheta)
\right\rangle
-
\frac12\vy^\top\vy.
\end{align}
Hence the FY conjugate update of the canonical parameters is the addition:
\begin{align}
\vlambda_{\mathcal D}
&= \vlambda_0 +
\begin{pmatrix}
\vX^\top\vy\\
-\frac12\vX^\top\vX
\end{pmatrix}
=
\begin{pmatrix}
\vX^\top\vy\\
-\frac12(\delta\vI+\vX^\top\vX)
\end{pmatrix}. \nonumber
\end{align}

In terms  $\vS$ and $\vm$, 
\begin{align}
\vlambda_{\mathcal D}
&=
\begin{pmatrix}
\vS\vm\\
-\frac12\vS
\end{pmatrix}, \nonumber \\
\left\langle \vlambda_{\mathcal D},\vT(\vtheta) \right\rangle
&= -\frac12 (\vtheta-\vm)^\top \vS (\vtheta-\vm) + \frac12\vm^\top\vS\vm. \nonumber
\end{align}
Whenever $\vlambda_{\mathcal D}$ belongs to the $q$-Gaussian canonical domain, the exact FY posterior is
\[
\pi_{\mathcal D}^\star(\vtheta)
= \exp_q
\left(
\left\langle
\vlambda_{\mathcal D},\vT(\vtheta)
\right\rangle
- A_{2-q}(\vlambda_{\mathcal D})
\right).
\]
The $q$-Gaussian posterior has location $\vm$, which coincides with the mean and MAP of classical ridge regression. 
The matrix $\vS$ is the canonical precision; $\vS^{-1}$ is equal to the posterior covariance in the Gaussian case, but not in general.
In \cref{app:reg} we derive the dispersion $\vSigma$, covariance, and support from $\vS$ for general $q$.

\subsection{Clustering with FYEM} \label{sec:gmm1}

\begin{figure*}[h] 
\centering
\includegraphics[width=\linewidth]{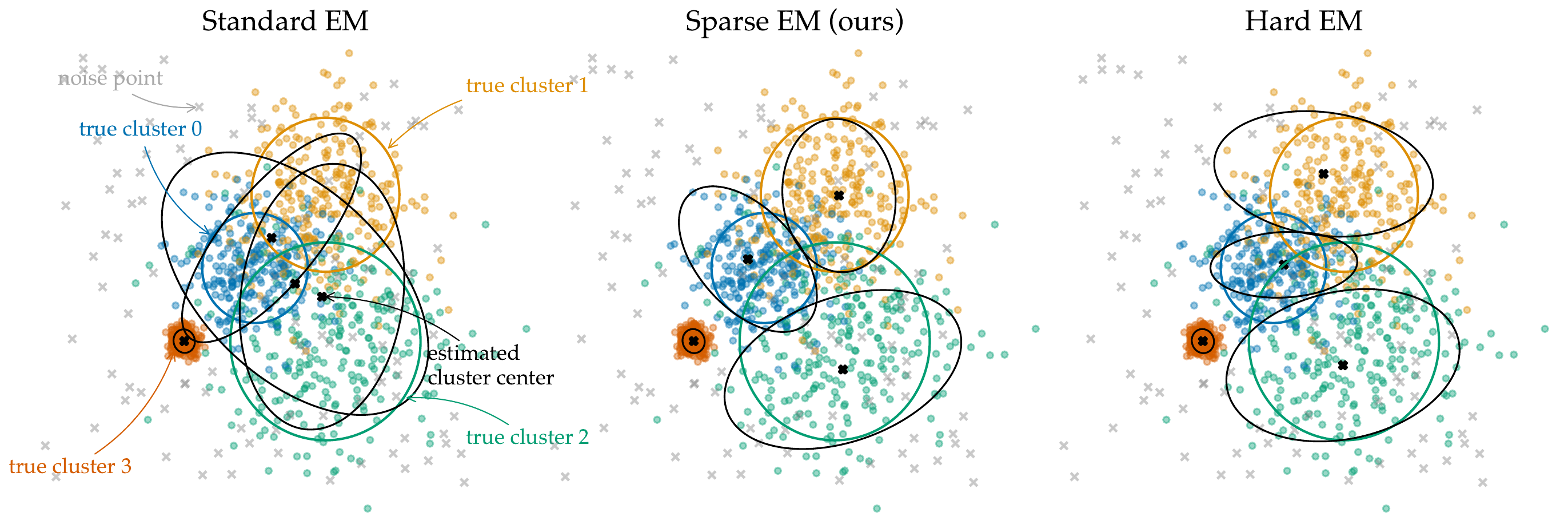}
\caption{
We compare standard, sparse, and hard EM on GMM-based clustering. Data points are colored by their true labels; \textcolor{gray}{\sf x} indicates noise points. Colored ellipses denote the data-generating covariances (the same across the three panels). Black ellipses are level curves of the estimated GMM components and illustrate the effect of E-step sparsity on the fit. Sparse EM is less influenced by the outliers while retaining soft assignments.
}
\label{fig:clustering_comparison}
\end{figure*}

Let $\vy_i\in\mathbb R^D$, and each \smash{$\theta_i\in\Theta=\{1,\ldots,K\}$} be its local unknown component label. Component parameters $\vphi_k$'s, and the mixing distribution $\pi_0$ are shared across observations and estimated in the M-step.
Let
\[
\pi_i = \left(\pi_i(1),\ldots,\pi_i(K)\right)
\in\Delta_K,
\ 
\pi=(\pi_1,\ldots,\pi_N)\in\Delta_K^N,
\]
where \( \pi_i(k) = \pi_i(\theta_i=k) \) is the local posterior mass assigned to component $k$ for observation $\vy_i$.

On a finite latent space, the component indicators span all functions of $\theta_i$. Any finite loss therefore has the affine form required for conjugacy. The prior potential vector, $\eta_0= (\eta_{0,1},\ldots,\eta_{0,K}) \in\mathbb R^K$, induces the mixing distribution $\pi_0\in\partial\Omega^*(\eta_0)$. The E-step adds the component log-likelihoods to these prior potentials and applies the induced posterior map
\[
\pi_0
= \left(\pi_0(1),\ldots,\pi_0(K)\right)
\in\Delta_K.
\]
Let \(\vphi = \{(\vmu_k,\vSigma_k)\}_{k=1}^K \) collect Gaussian component parameters, so that
\[
P(\theta_i=k) = \pi_0(k),
\quad
p_{\text{\vphi}}(\vy_i\mid\theta_i=k) = \mathcal N(\vy_i;\vmu_k,\vSigma_k).
\]

The FYEM criterion with the logarithmic observation loss
\smash{\( 
\ell_{\text{\vphi}}(\vy_i;k)
= -\log \mathcal N(\vy_i;\vmu_k,\vSigma_k)
\)}, is
\begin{align}
&F_{\mathcal D}(\pi,\vphi,\eta_0) = \nonumber \\
&\sum_{i=1}^N
\left\{ -\sum_{k=1}^K \pi_i(k)
\log \mathcal N(\vy_i;\vmu_k,\vSigma_k)
+ \mathbb D_\Omega(\eta_0;\pi_i)
\right\}. 
\end{align}

Given $(\vphi^{(r)},\eta_0^{(r)})$, the unrestricted E-step is
\begin{align}
\pi_i^{(r+1)}\!\in\!\partial\Omega^*
\left[\left(\eta_{0,k}^{(r)}+\log \mathcal{N}\left(\vy_i;\vmu_k^{(r)},\vSigma_k^{(r)}\right)\right) \right]. 
\label{eq:gmm_fy_estep}
\end{align}
The M-step, standard and hard E-steps are in \cref{app:gmm-updates}.

\paragraph{Sparse FYEM.}
Under the Tsallis $\Omega_{2-q}$ negentropy, the potential representing the current mixing distribution satisfies
\[
\eta_{0,k}^{(r)}
= \frac{\left( \pi_0^{(r)}(k) \right)^{1-q}}{1-q} + C,
\qquad
k=1,\ldots,K,
\]
where $C$ is a constant common to all coordinates. Substituting this potential into \cref{eq:gmm_fy_estep}, and dropping $C$ by shift invariance, we get
\begin{align}
&\pi_i^{(r+1)} = \nonumber \\
& \ \ (2-q)\mbox{-entmax}
\left(\left( \eta_{0,k}^{(r)} + \log \mathcal N \left( \vy_i;\vmu_k^{(r)},\vSigma_k^{(r)} \right) \right)_{k=1}^K \right). \nonumber 
\label{eq:sparseEstep}
\end{align}
The limit $q\to1$ recovers the EM's softmax E-step.

\begin{table}[t!]
\caption{
Clustering quality metrics (higher is better), for GMM clustering using standard EM, hard EM, and sparse FYEM with $q=0$. The mixing distribution $\pi_0$, and the Gaussian component parameters are learned with the shared M-step. Values are averaged over 5 random seeds $\pm$ standard errors.
}
\label{tab:synthetic-clustering} 
\begin{center}
\vspace{0.3cm}
\begin{tabular}{p{1.3cm}p{1.8cm}p{1.8cm}p{1.8cm}}
\toprule
\textsc{E-Step} & \textsc{Adj. MI} & \textsc{Adj. RI} & \textsc{Silhouette} \\
\midrule
Standard & $.606 \pm .006$ & $0.531 \pm .002$ & $.345 \pm .016$ \\
Hard & $.537 \pm .044$ & $0.348 \pm .022$ & $.207 \pm .064$ \\
Sparse & $\textbf{.670} \pm .008$ & $\textbf{.668} \pm .012$ & $\textbf{.442} \pm .007$ \\
\bottomrule
\end{tabular}
\end{center}
\end{table}

When $q<1$, local posteriors $\pi_i^{(r+1)}$'s may contain both zero and positive coordinates. Experiment details and sensitivity to $q$ are described in \cref{ssec:clustering}. In particular, $\pi_i^{(r+1)}(k)=0$ means that observation $\vy_i$ has zero weight in the M-step update of component $k$. Standard EM uses dense local posteriors whose mixing probabilities are all positive. Hard EM keeps only one component for each observation. Sparse FYEM can exclude some components while retaining a soft posterior over the remaining components. For $q<1$, the vectors $\pi_i$ are generalized FY posteriors. The sparsity in question concerns posterior support, rather than zero-valued parameter estimates.\footnote{
FYEM, with Tsallis $(2-q)$-negentropy regularization ($q < 1$) characterizes qualitatively different sparsity, compared to sparsity-inducing priors, such as LASSO (equivalently, Laplace priors \citep{LASSO, Hastie}), or spike-and-slab priors \citep{spike}. Sparsity-inducing priors encourage the estimated object (e.g. coefficients of parameters) to be sparse, but the associated posterior distributions have full support. For example, the Bayes posterior of a Laplace sparsity-inducing prior and a Gaussian observation model has full support. Such priors are said to induce sparsity because the posterior mode (MAP estimate) can be exactly zero. In contrast, using non-sparsity-inducing Gaussian priors and a Gaussian observation models in FY Bayes with Tsallis $2$-negentropy regularization, to get bounded-support generalized posteriors. A different use of ``sparse'' appears in the EM algorithm of \citet[Sec.~5]{neal1998view}, who use the Shannon free energy and reduce computation by freezing small responsibilities between occasional full updates, rather than setting them to zero. 
}
Sparse FYEM has the highest scores on three clustering metrics (\cref{tab:synthetic-clustering}). \cref{fig:clustering_comparison} illustrates the component fits, with sparse FYEM less influenced by outliers.

\subsection{Fenchel-Young $\beta$-VAEs}\label{sec:FYVAE_beta}


In \citet{kingma2013auto}'s VAE a Gaussian variational posterior and a $\mathcal{N}(\vzero, \vI)$ prior are combined with a product of independent Bernoulli's observation model, representing the pixels of MNIST images.
\cref{ssec:vaes} reviews the baseline objective and training details.

\begin{table}[h]
\centering
\caption{VAE test $l_1$ reconstruction error: mean $\pm$ std.
error over 5 seeds. The latent variable index $q$ selects Gaussian ($q=1$), biweight ($q=0.5$), or Epanechnikov ($q=0$) posteriors.
Observation models are Bernoulli ($q'=1$) or sparse binary ($q'=0$).}
\label{tab:beta-Gaussian-vae}
\begin{tabular}{@{}ccrr@{}}
\toprule
\shortstack{$q'$} &
\shortstack{$q$} &
\textsc{MNIST} &
\shortstack{\textsc{Fashion MNIST}} \\
\midrule
1 & 1   & $13.236\pm.004$ & $34.049\pm.019$ \\
1 & 0.5 & $12.065\pm.021$ & $32.480\pm.018$ \\
1 & 0   & $12.086\pm.012$ & $32.537\pm.025$ \\
\midrule
0 & 1   & $12.567\pm.017$ & $32.881\pm.027$ \\
0 & 0.5 & $9.261\pm.009$  & $25.970\pm.086$ \\
0 & 0   & $\mathbf{9.144}\pm.020$
        & $\mathbf{25.851}\pm.072$ \\
\bottomrule
\end{tabular}
\end{table}

We experiment with bounded support $q$-Gaussian\footnote{\citet{martins2022sparse} call this family $\beta$-Gaussian.} variational posteriors. 
We also compare the baseline product of Bernoulli's decoder with a product of binary deformed exponential family distributions induced by another convex generator $\Omega'_{2-q'}$. 
Here $\Omega_{2-q}$ acts on continuous latent densities, whereas $\Omega'_{2-q'}$ acts on discrete probability mass functions over the two outcomes of each pixel. 
The FYELBO objective is given as follows:
\begin{equation}
\sum\limits_{i=1}^N
\underset{\text{\vtheta}_i\sim\pi_i}{\mathbb E}
\left[
\ell_{\text{\vphi},\Omega'_{2-q'}}(\vx_i;\vtheta_i)
\right]
+ \beta
\mathbb D_{\Omega_{2-q}}
\left( \eta_0;\pi_i \right).
\end{equation}
Here, \( \pi_i = \pi_{\text{\vpsi}}(\cdot\mid \vx_i), \) where $\vpsi$ denotes encoder parameters and $\vphi$ denotes decoder parameters. 
We fix the prior potential \( \eta_0(\vtheta)=-\frac12\|\vtheta\|_2^2 \), to represent a standard Gaussian prior under Shannon geometry, and a $q$-Gaussian prior under Tsallis geometry. 
We use $q$-Gaussian variational posteriors $\pi_{\text{\vpsi}}(\cdot\mid \vx_i)$.

In \cref{tab:beta-Gaussian-vae} we report the results of computationally efficient combinations of $\Omega'$ and $\Omega$ in VAEs trained on MNIST and FashionMNIST. 
We find the Epanechnikov ($q=0$) posterior to give overall lowest reconstruction error, which was also observed by ~\citep{qin2024kernel}.\footnote{
\citet{qin2024kernel} construct kernel-based variational posteriors and replace the ELBO's KL term with an asymptotically motivated surrogate. 
Their Epanechnikov kernel minimizes this surrogate at fixed kernel mean and variance, equivalently maximizing Tsallis $2$-entropy.
We instead define the posterior by unrestricted minimization of the full FY objective (\cref{eq:opt_post}), with Tsallis negentropy as one
choice of generator.
}
We note that sparsity in the encoder and in the decoder separately contribute to improved performance. We report average $l_1$ as a common
reconstruction criterion, since the FYELBO values themselves are not directly comparable. 
We tune $\beta$ by cross-validation to prevent the posterior from collapsing to the prior. 


\section{Conclusion}

Fenchel-Young Bayes expands the class of variational posteriors, while preserving the Bayesian coupling of prior potentials, discrepancies and posteriors through a convex generator function. Further, we define FY evidence and its variational lower bound, derive FYEM and amortized FY latent-variable learning; we illustrate FYEM with sparse Gaussian mixtures, and show small FYVAEs have lower image reconstruction errors than the corresponding VAE baselines.
We characterize posterior updating as a translation by the loss where, affine potential families let us generalize conjugate updating; we illustrate exact conjugate updating with bounded-support and heavy-tailed regression posteriors.


\subsection*{AI use statement}


In this work, we used generative AI tools for figure and table formatting, literature search, proofreading and revising.
We have not used generative AI tools for research ideas, code, or writing. Nothing else is applicable to this work. 
We have reviewed all AI-assisted work.
We take responsibility for the final content of this work, including text, claims, or artifacts produced with the aid of generative AI.



\bibliography{clean_references}

\clearpage
\appendix
\thispagestyle{empty}

\onecolumn














\aistatstitle{Supplementary Materials}

\section{Variational Bayes and Conjugate Updating}
\label[appendix]{appendix:vcb}

\paragraph{Variational Bayes.}
Let the prior and candidate posterior densities on $\Theta$ be defined with respect to a fixed reference measure $h_\Theta$, and define
\[
\ell_{\mathcal D}(\vtheta)
= -\sum_{i=1}^N\log p(\mathcal D_i\mid\vtheta).
\]
When $0<\int_\Theta e^{-\ell_{\mathcal D}(\vtheta)}\pi_0(\vtheta)<\infty$,
the Bayes posterior is the unique unrestricted minimizer
\begin{align}
\pi_{\mathcal D}^\star
=\arg\min_{\pi\in\mathcal P} \mathbb E_\pi[\ell_{\mathcal D}(\vtheta)] + \mathbb D_{\rm KL}(\pi\|\pi_0) 
= \arg\min_{\pi\in\mathcal P} \mathbb E_\pi[\ell_{\mathcal D}(\vtheta)-\log\pi_0(\vtheta)] - \mathcal H(\pi).
\end{align}
Here $\mathcal H$ is the Shannon entropy relative to $h_\Theta$, and the KL divergence is $+\infty$ if $\pi$ is not absolutely continuous with respect to $\pi_0$.

For a general measurable loss $\ell$, let
\[
Z = \int_\Theta e^{-\ell(\text{\vtheta})}\pi_0(\vtheta),
\qquad
\pi_\ell^\star(\vtheta) 
= \frac{e^{-\ell(\text{\vtheta})}\pi_0(\vtheta)}{Z}.
\]
Whenever this density is well defined,
\begin{equation}
\mathbb E_\pi[\ell(\vtheta)] + \mathbb D_{\rm KL}(\pi\|\pi_0)
= \mathbb D_{\rm KL}(\pi\|\pi_\ell^\star) - \log Z.
\label{eq:gibbs_decomposition}
\end{equation}
Thus, before any parametric family assumption, $\pi_\ell^\star$ is an exact unrestricted optimizer: it is the standard Bayes posterior when $\ell$ is a negative log likelihood, and the corresponding Gibbs posterior when $\ell$ is a more general loss.

We derive finite dimensional exponential family posteriors by restricting the unnormalized log-density score to an affine function of a finite statistic. For statistic $\vT:\Theta\to\mathbb R^P$, fixed base measure $b$, and $\vlambda$ such that $A (\vlambda)<\infty$, let
\begin{equation}
\pi_{\text{\vlambda}}(\vtheta) = b(\vtheta) \exp\left(\langle\vlambda,\vT(\vtheta)\rangle -A(\vlambda) \right),
\qquad
A(\vlambda) = \log\int_\Theta b(\vtheta)e^{\langle \text{\vlambda}, \text{\vT}(\text{\vtheta})\rangle} .
\label{eq:posterior_expfam}
\end{equation}
When $b$ is constant, $\pi_{\text{\vlambda}}$ maximizes Shannon entropy among all densities with the same moments
$\mathbb E_\pi[\vT(\vtheta)] =\mathbb E_{\pi_{\text{\vlambda}}}[\vT(\vtheta)]$.
When $b$ is not constant, an analogous characterization is possible in terms of the relative entropy with respect to the measure $b \ dh_\Theta$.

\paragraph{Conjugate Bayes.}
Suppose the prior $\pi_{\text{\vlambda}_0}$ has the exponential family form defined in \cref{eq:posterior_expfam}. Assume that the
likelihood factorizes over observations and that each log-likelihood
term is affine in $\vT(\vtheta)$:
\begin{equation}
\log p(\mathcal D_i\mid\vtheta)
= \langle \vt(\mathcal D_i), \vT(\vtheta) \rangle - c(\mathcal D_i).
\end{equation}
For a sampling model written in canonical exponential family form, $\vT(\vtheta)$ contains all $\vtheta$-dependent terms, including both the natural or canonical parameter and the negative sampling log-partition. 
To use this form, we take $\vt(\mathcal D_i)\in\mathbb R^P$ to comprise both the usual data-dependent sufficient statistic and the coefficient of that log-partition term, and view it as a vector of data-dependent coefficients.
The scalar function $c(\mathcal D_i)$ does not depend on $\vtheta$. 

According to standard Bayes rule
\begin{align}
\pi(\vtheta\mid\mathcal D)
&= b(\vtheta) \exp\left( \langle \vlambda_{\mathcal D}, \vT(\vtheta) \rangle - A(\vlambda_{\mathcal D}) \right),
\label{eq:posterioref}
\end{align} 
using the additive log score space update:
\begin{align} \label{eq:addthelambdas}
\vlambda_{\mathcal D}
&= \vlambda_0 + \sum_{i=1}^N \vt(\mathcal D_i).
\end{align}
The posterior stays in the same family as the prior because the data change only the canonical coefficients defining the affine function of $\vT(\vtheta)$, which we will call \textbf{scoring function}, $\eta_{\text{\vlambda}}$. The update is well defined when $\vlambda_{\mathcal D}$ belongs to the canonical parameter domain
\[
\Lambda = \left\{ \vlambda\in\mathbb R^P: A(\vlambda)<\infty \right\}.
\]
The family $\{\pi_{\text{\vlambda}}:\vlambda\in\Lambda\}$ is conjugate to a class of likelihood functions when every likelihood contributes a coefficient increment in the same score basis $\vT$, and $\vlambda_0+\vt(\mathcal D)\in\Lambda$ for every $\vlambda_0\in\Lambda$ and every dataset $\mathcal D$ in the class.\footnote{Since the log-partition domain is convex, the whole path \(\vlambda_0+\omega\vt(\mathcal D)\), \(0\leq\omega\leq1\),
also stays in the family whenever its endpoint does.}
This is the standard exponential-family conjugate update; see \citet{Barndorff-Nielsen} and \citet{diaconis1979conjugate}.

\section{Additional Derivations}
\label[appendix]{app:derivations}

\subsection{Maximum Entropy Characterization}\label[appendix]{app:maxent}

\begin{prop}
Assume that for $\vlambda \in \real^P$ the supremum defining $\Omega^*(\eta_{\text{\vlambda}})$ is attained by $\pi_{\text{\vlambda}} \in \mathcal P$:
\(
\Omega^*(\eta_{\text{\vlambda}}) = \mathbb E_{\pi_{\text{\vlambda}}}[\eta_{\text{\vlambda}}(\vtheta)] - \Omega(\pi_{\text{\vlambda}})
\).
Define
\(
\vmu(\vlambda) = \mathbb E_{\pi_{\text{\vlambda}}}[\vT(\vtheta)]
\).
Then $\pi_{\text{\vlambda}}$ minimizes $\Omega$ over all $\pi\in\mathcal P$ satisfying $\mathbb E_\pi[\vT(\vtheta)]=\vmu(\vlambda)$.
\end{prop}

\begin{proof}
By the optimality of $\pi_{\text{\vlambda}}$ in the definition of $\Omega^*(\eta_{\text{\vlambda}})$, for every \smash{$\pi \in \mathcal P$},
\smash{\(
\Omega(\pi) - \mathbb E_{\pi}[\eta_{\text{\vlambda}}(\vtheta)]
\ge \Omega(\pi_{\text{\vlambda}}) - \mathbb E_{\pi_{\text{\vlambda}}}[\eta_{\text{\vlambda}}(\vtheta)]
\)}.
Let $\pi$ satisfy the same moment constraints as $\pi_{\text{\vlambda}}$, namely $\mathbb E_{\pi}[\vT (\vtheta)] = \vmu(\vlambda)$. Then
\begin{align}
\mathbb E_{\pi}[\eta_{\text{\vlambda}}(\vtheta)]
&= \langle \vlambda, \mathbb E_{\pi}[\vT(\vtheta)] \rangle 
= \langle \vlambda, \mathbb E_{\pi_{\text{\vlambda}}}[\vT(\vtheta)] \rangle 
= \mathbb E_{\pi_{\text{\vlambda}}}[\eta_{\text{\vlambda}}(\vtheta)]. 
\end{align}
Hence $\Omega(\pi) \ge \Omega(\pi_{\text{\vlambda}})$, which proves optimality. If $\Omega$ is strictly convex, this minimizer is unique.
\end{proof}

\subsubsection{$q$-Exponential Family as Max Tsallis $(2-q)$-Entropy}\label[appendix]{ssec:tsallis_maxent}

Let $(\Theta, \mathcal{A}, h_\Theta)$ be a measure space, and let
\[
\vT(\vtheta) = \left(T_1(\vtheta), ..., T_P(\vtheta) \right)
\]
be measurable. Fix $q < 2$, $q \neq 1$, and define
\[
\beta_{q}(t)
= \begin{cases}
\displaystyle
\frac{t^{2-q}-t}{(2-q)(1-q)},
& t \ge 0, \\
+ \infty, & t < 0
\end{cases}
\]
and
\[
\Omega_{2-q}(\pi) = \int_\Theta \beta_{q} (\pi(\vtheta)) dh_\Theta(\vtheta).
\]
On the simplex of densities,
\[
\Omega_{2-q}(\pi) = -\frac{1}{2-q}\mathcal S_{2-q}(\pi),
\qquad
\mathcal{S}_{2-q}(\pi) = \frac{1 - \int_\Theta \pi(\vtheta)^{2-q} d h_\Theta(\vtheta)}{1-q}.
\]

The deformed log, which is the inverse of $\exp_q$, is defined as 
\[
    \log_{q}(u) = 
    \left\{
    \begin{array}{ll}
        \frac{u^{1-q}-1}{1-q} & \text{if $q \ne 1$} \\
        \log u & \text{if $q=1$}
    \end{array}
    \right. . 
\]

\begin{prop}
Assume there exist $\vmu = (\mu_1, ...,\mu_P)$ and $\vlambda = (\lambda_1, ..., \lambda_P)$ such that
\[
\pi_{\text{\vlambda}} (\vtheta) =
\exp_{q} \left(\langle \vlambda,\vT(\vtheta) \rangle - A_{2-q}(\vlambda) \right)
\]
is a density, $\Omega_{2-q}(\pi_{\text{\vlambda}}) < \infty$, all the integrals below are finite, and
\[
\mu_i = \int_\Theta T_i(\vtheta) \pi_{\text{\vlambda}}(\vtheta) dh_\Theta(\vtheta),
\qquad i = 1, ..., P.
\]
Then $\pi_{\text{\vlambda}}$ attains $\Omega^*_{2-q}(\eta_{\text{\vlambda}})$. Consequently, it minimizes $\Omega_{2-q}$, equivalently maximizes
$\mathcal S_{2-q}$, over all densities satisfying
\[
\mu_i = \int_\Theta T_i(\vtheta) \pi(\vtheta) dh_\Theta(\vtheta),
\qquad i = 1, ..., P.
\]

\end{prop}

\begin{proof}
Set
\[
\tilde\eta_{\text{\vlambda}}(\vtheta) = \langle \vlambda, \vT(\vtheta) \rangle - A_{2-q}(\vlambda).
\]

We claim that for $h_\Theta$-a.e. $\vtheta$,
\[
\tilde\eta_{\text{\vlambda}}(\vtheta) + \frac{1}{2-q} \in \partial \beta_{q} \left(\pi_{\text{\vlambda}} (\vtheta) \right).
\]

If $\pi_{\text{\vlambda}}(\vtheta) > 0$, then
\( 
\log_{q}\pi_{\text{\vlambda}} (\vtheta) = \tilde\eta_{\text{\vlambda}} (\vtheta),
\)
and since
\[
\beta_{q}'(u) = \log_{q}(u) + \frac{1}{2-q}, \qquad u > 0,
\]
it follows that
\[
\tilde\eta_{\text{\vlambda}}(\vtheta) + \frac{1}{2-q} = \beta'_{q} \left(\pi_{\text{\vlambda}}(\vtheta) \right) \in \partial \beta_{q} \left(\pi_{\text{\vlambda}} (\vtheta) \right).
\]

If $q < 1$ and $\pi_{\text{\vlambda}} (\vtheta) = 0$, then by the definition of $\exp_{q}$ we have
\[
\tilde\eta_{\text{\vlambda}}(\vtheta) \le -\frac{1}{1-q}.
\]
Also, the right derivative at 0 is
\[
\beta'_{q}(0) = -\frac{1}{(1-q) (2-q)},
\qquad 
\partial \beta_{q}(0) = (-\infty, \beta'_{q}(0)].
\]
Therefore
\[
\tilde\eta_{\text{\vlambda}}(\vtheta)+\frac{1}{2-q} \le - \frac{1}{1-q} + \frac{1}{2-q}
= -\frac{1}{(1-q)(2-q)}
= \beta'_{q}(0),
\]
and again
\[
\tilde\eta_{\text{\vlambda}}(\vtheta) + \frac{1}{2-q} \in \partial \beta_{q} \left(\pi_{\text{\vlambda}}(\vtheta)\right).
\]



Let $\pi\in\mathcal P$ be arbitrary, with the displayed quantities finite. Integrating the pointwise subgradient inequality 
\begin{align*}
\Omega_{2-q}(\pi)-\Omega_{2-q}(\pi_{\text{\vlambda}})
&\geq \int_\Theta
\left(
\tilde\eta_{\text{\vlambda}}(\vtheta)+\frac{1}{2-q}
\right)
\left(\pi(\vtheta)-\pi_{\text{\vlambda}}(\vtheta)\right) dh_\Theta(\vtheta)\\
&= \mathbb E_\pi[\eta_{\text{\vlambda}}(\vtheta)] - \mathbb E_{\pi_{\text{\vlambda}}} [\eta_{\text{\vlambda}}(\vtheta)].
\end{align*}
The constant terms cancel because both densities integrate to one. Rearranging 
\[
\mathbb E_\pi[\eta_{\text{\vlambda}}(\vtheta)]
-\Omega_{2-q}(\pi)
\leq
\mathbb E_{\pi_{\text{\vlambda}}}
[\eta_{\text{\vlambda}}(\vtheta)]
-\Omega_{2-q}(\pi_{\text{\vlambda}}).
\]
Thus no candidate gives a conjugate objective with larger value than that of $\pi_{\text{\vlambda}}$. Since equality holds
at $\pi=\pi_{\text{\vlambda}}$,
\[
\Omega^*_{2-q}(\eta_{\text{\vlambda}})
= \mathbb E_{\pi_{\text{\vlambda}}}
[\eta_{\text{\vlambda}}(\vtheta)] - \Omega_{2-q}(\pi_{\text{\vlambda}}).
\]

Now restrict $\pi$ to have the same moments of $\vT$ as $\pi_{\text{\vlambda}}$. Since $\eta_{\text{\vlambda}}=\langle\vlambda,\vT\rangle$,
\[
\mathbb E_\pi[\eta_{\text{\vlambda}}(\vtheta)]
=
\langle\vlambda,\vmu\rangle
=
\mathbb E_{\pi_{\text{\vlambda}}}
[\eta_{\text{\vlambda}}(\vtheta)].
\]
The potential expectations cancel, leaving $\Omega_{2-q}(\pi)\geq
\Omega_{2-q}(\pi_{\text{\vlambda}})$.
Since $\Omega_{2-q}=-\mathcal S_{2-q}/(2-q)$ and $q<2$, this proves the maximum Tsallis entropy claim.
\end{proof}

\subsection{Conjugate FY Linear Regression}
\label[appendix]{app:reg}

\paragraph{Canonical form.}
With $\delta>0$, the regression update is
\[
\vS=\delta\vI+\vX^\top\vX,
\qquad
\vm=\vS^{-1}\vX^\top\vy.
\]
The matrix $\vS$ is positive definite. We assume
\[
q<\frac{D+4}{D+2}, \qquad q\neq1,
\]
which ensures finite second moments for the densities below.

From the completed quadratic score 
\[
\pi_{\mathcal D}^\star(\vtheta)
=
\exp_q\left(
-\frac12(\vtheta-\vm)^\top\vS(\vtheta-\vm)
+\frac12\vm^\top\vS\vm
-A_{2-q}(\vlambda_{\mathcal D})
\right).
\]
For $q<1$, let $\gamma=D+2/(1-q)$; for $q>1$, let $\nu=2/(q-1)-D$. We first express the unknown normalizing constant in terms of $R(q,D)>0$ or $\kappa_t(\nu,D)>0$, which we determine below
\begin{align}
A_{2-q}(\vlambda_{\mathcal D})
&=
\frac12\vm^\top\vS\vm
+
\begin{cases}
\displaystyle
-\frac12\left(
R(q,D)^2|\vS|^{1/\gamma}-\frac{2}{1-q}
\right),
& q<1,\\[2mm]
\displaystyle
\frac{1}{q-1}\left(
|\vS|^{-1/\nu}\frac{\nu}{\nu+D}\kappa_t(\nu,D)-1
\right),
& q>1.
\end{cases}
\label{eq:A_regression_cases}
\end{align}

\paragraph{Radial representation.}
For a positive definite dispersion matrix $\vSigma$, define the centered, rescaled vector $\vz$ and its radius $r$ by
\begin{align}
\vz=\vSigma^{-1/2}(\vtheta-\vm)=r\vu,
\qquad
r^2(\vtheta)
=(\vtheta-\vm)^\top\vSigma^{-1}(\vtheta-\vm),
\qquad
\vtheta=\vm+\vSigma^{1/2}(r\vu).
\label{eq:polar}
\end{align}
For the elliptical densities below, $\vu$ is uniform on $S^{D-1}=\{\vu:\vu^\top\vu=1\}$, and $r\geq0$ is independent of $\vu$.

\paragraph{$q<1$.}
Let
\[
\vSigma=|\vS|^{1/\gamma}\vS^{-1},
\qquad R=R(q,D).
\]
Substituting the $q<1$ expression in \cref{eq:A_regression_cases} into the canonical density and expanding $\exp_q$ 
\begin{align*}
\pi_{\mathcal D}^\star(\vtheta)
&=
\left[
\frac{1-q}{2}
\left(
R^2|\vS|^{1/\gamma}
-(\vtheta-\vm)^\top\vS(\vtheta-\vm)
\right)
\right]_+^{\frac{1}{1-q}}\\
&=
|\vSigma|^{-1/2}
\left[
\frac{1-q}{2}\left(R^2-r^2(\vtheta)\right)
\right]_+^{\frac{1}{1-q}}.
\end{align*}
The second equality uses $\vS=|\vS|^{1/\gamma}\vSigma^{-1}$ and $|\vSigma|^{-1/2}=|\vS|^{1/(\gamma(1-q))}$.

Under the change of variables in \cref{eq:polar}, the Jacobian cancels $|\vSigma|^{-1/2}$. Normalization therefore requires
\[
\int_{\real^D}
\left[
\frac{1-q}{2}\left(R^2-\|\vz\|^2\right)
\right]_+^{\frac{1}{1-q}}
\,d\vz=1.
\]
Integrating in polar coordinates and changing the radial integration variable from \(r\) to \(r^2/R^2\) it follows that 
\[
R(q,D)
=
\left[
\frac{\Gamma\left(\frac D2+\frac{2-q}{1-q}\right)}{\pi^{D/2}\Gamma\left(\frac{2-q}{1-q}\right) }
\left(\frac{2}{1-q}\right)^{\frac{1}{1-q}}
\right]^{\frac{1-q}{2+D(1-q)}},
\]
where $\Gamma$ is the Gamma function. The posterior support is
\[
\operatorname{supp}\left(\pi_{\mathcal D}^\star\right)
=
\left\{
\vtheta:r^2(\vtheta)\leq R(q,D)^2
\right\}.
\]
To obtain the posterior covariance, we need $\mathbb E[r^2]$. The same change of variables gives identifies the squared radial fraction:
\begin{align}
\frac{r^2}{R(q,D)^2}
\sim
\mathrm{Beta}\left(\frac D2,\frac{2-q}{1-q}\right),
\qquad r\in[0,R(q,D)].
\label{eq:radial_bounded}
\end{align}

\paragraph{$q>1$.}
We will use the dispersion matrix
\[
\vSigma = \kappa_t(\nu,D)|\vS|^{-1/\nu}\vS^{-1}.
\]
Substituting the $q>1$ expression in \cref{eq:A_regression_cases} and using $1/(q-1)=(\nu+D)/2$ 
\begin{align}
\pi_{\mathcal D}^\star(\vtheta)
&=
\left[
|\vS|^{-1/\nu}\frac{\nu}{\nu+D}\kappa_t(\nu,D)
+\frac{q-1}{2}
(\vtheta-\vm)^\top\vS(\vtheta-\vm)
\right]^{-\frac{1}{q-1}}
\nonumber\\
&\propto
\left(1+\frac{r^2(\vtheta)}{\nu}\right)^{-(\nu+D)/2}.
\label{eq:pearsonVII_S_form}
\end{align}
This is the Student-$t$ kernel. Its normalized density is
\[
\mathcal T_\nu(\vtheta\mid\vm,\vSigma)
=
\frac{1}{Z_\nu|\vSigma|^{1/2}}
\left(1+\frac{r^2(\vtheta)}{\nu}\right)^{-(\nu+D)/2},
\qquad
Z_\nu =\frac{\Gamma(\nu/2)(\nu\pi)^{D/2}}{\Gamma((\nu+D)/2)}.
\]
Normalization requires the multiplicative normalization constant in the canonical density to equal $1/(Z_\nu|\vSigma|^{1/2})$.
Substituting the definition of $\vSigma$ and solving for
$\kappa_t$ we get
\[
\kappa_t(\nu,D)
=
\frac{\nu+D}{\nu}
\left[
(\nu+D)^{D/2}\pi^{D/2}
\frac{\Gamma(\nu/2)}{\Gamma((\nu+D)/2)}
\right]^{2/\nu}.
\]
Thus
$\pi_{\mathcal D}^\star(\vtheta)
=\mathcal T_\nu(\vtheta\mid\vm,\vSigma)$.
The density is proper for $\nu>0$; the range we assume for values of $q$ ensures $\nu>2$.

Using $\vz=r\vu$ and integrating out the angular part,
the radius has density proportional to
\[
r^{D-1}
\left(1+\frac{r^2}{\nu}\right)^{-(\nu+D)/2},
\qquad r\geq0.
\]
Changing variables from $r$ to $r^2$ gives the kernel
\[
(r^2)^{D/2-1}
\left(1+\frac{r^2}{\nu}\right)^{-(\nu+D)/2}.
\]
Hence
\[
\frac{r^2}{\nu}
\sim
\mathrm{BetaPrime}\left(\frac D2,\frac{\nu}{2}\right).
\]
Equivalently,
\begin{align}
r^2
\stackrel{d}{=}
\nu\frac{\chi_D^2}{\chi_\nu^2}
\stackrel{d}{=}
D F_{D,\nu},
\label{eq:radial_t}
\end{align}
where the chi-squared variables are independent.

\paragraph{Covariance.}
By spherical symmetry, the posterior mean is $\vm$. Independence of $r$ and $\vu$, and the fact that $\mathbb E[\vu\vu^\top]=\vI/D$ imply
\begin{align}
\Var_{\pi_{\mathcal D}^\star}(\vtheta)
&=
\vSigma^{1/2}\mathbb E[r^2]
\mathbb E[\vu\vu^\top]\vSigma^{1/2}
=
\frac{\mathbb E[r^2]}{D}\vSigma.
\label{eq:elliptical_var}
\end{align}
Using the radial laws in \cref{eq:radial_bounded,eq:radial_t}, we obtain
\[
\Var_{\pi_{\mathcal D}^\star}(\vtheta)
=
\begin{cases}
\kappa(q,D)\vSigma, & q<1,\\[1mm]
\displaystyle\frac{\nu}{\nu-2}\vSigma, & q>1,
\end{cases}
\qquad
\kappa(q,D)
=
\frac{R(q,D)^2}{D+\frac{2(2-q)}{1-q}}.
\]

\subsection{GMM Updates}
\label[appendix]{app:gmm-updates}

\paragraph{Shared M-step.}
Given the local posterior reports $\pi^{(r+1)}$, define the effective component count
\[
N_k^{(r+1)} = \sum_{i=1}^N \pi_i^{(r+1)}(k).
\]
For $N_k^{(r+1)}>0$, the Gaussian parameter updates are
\begin{align}
\vmu_k^{(r+1)} &= \frac{1}{N_k^{(r+1)}} \sum_{i=1}^N \pi_i^{(r+1)}(k) \vy_i, \\
\vSigma_k^{(r+1)} &= \frac{1}{N_k^{(r+1)}}
\sum_{i=1}^N \pi_i^{(r+1)}(k)
\left( \vy_i-\vmu_k^{(r+1)} \right)
\left( \vy_i-\vmu_k^{(r+1)} \right)^\top.
\end{align}
For the unrestricted categorical latent marginal block, the mixing distribution is updated by
\begin{align}
\pi_0^{(r+1)}(k)
&= \frac{N_k^{(r+1)}}{N},
\qquad k=1,\ldots,K,
\label{eq:gmm_mixing_update}
\end{align}
or equivalently
\[
\pi_0^{(r+1)}
= \frac1N \sum_{i=1}^N \pi_i^{(r+1)}.
\]
The updated reference score is any score satisfying
\[
\pi_0^{(r+1)}
\in
\partial\Omega^*
\left(
\eta_0^{(r+1)}
\right).
\]
Standard EM, hard EM, and sparse FYEM use this same M-step and differ only in the E-step local posteriors. 
These updates depend on the local reports only through
\[
N_k^{(r+1)},\qquad
\sum_{i=1}^N\pi_i^{(r+1)}(k)\vy_i,\qquad
\sum_{i=1}^N\pi_i^{(r+1)}(k)\vy_i\vy_i^\top.
\]
These are the expected complete-data sufficient statistics: component counts, sums, and second moments. Sparse FYEM changes their weights, not the form of the M-step.

\paragraph{Standard EM.}
For the Shannon negentropy $\Omega_1=-\mathcal H$, the score representing the current mixing distribution is
\[
\eta_{0,k}^{(r)}
= \log\pi_0^{(r)}(k)+C,
\qquad
k=1,\ldots,K,
\]
where $C$ is common to all coordinates. Substituting this score into
\cref{eq:gmm_fy_estep} gives
\begin{align}
\pi_i^{(r+1)}(k)
&= \frac{ \pi_0^{(r)}(k) \mathcal N \left(\vy_i;\vmu_k^{(r)},\vSigma_k^{(r)} \right)}{\sum_{j=1}^K \pi_0^{(r)}(j) \mathcal N \left( \vy_i;\vmu_j^{(r)},\vSigma_j^{(r)} \right) },
\qquad
k=1,\ldots,K.
\label{eq:standard_gmm_estep}
\end{align}
Thus, in the Shannon case, $\pi_i^{(r+1)}$ is the ordinary
conditional distribution of the component label under the current
GMM. Together with the shared M-step above, this recovers standard
EM \citep{Figueiredo2002}.

\paragraph{Hard EM.}
Hard EM uses the same logarithmic component scores as standard EM,
but restricts each local posterior report to a vertex of the simplex:
\[
\pi_i^{(r+1)} \in
\left\{
\ve_k: k \in \arg\max_{1\leq j\leq K}
\left[
\log\pi_0^{(r)}(j) + \log \mathcal N
\left( \vy_i;\vmu_j^{(r)},\vSigma_j^{(r)}
\right)
\right]
\right\}.
\]
Thus each observation contributes to exactly one component update. This is the classification EM algorithm \citep{Celeux}; its M-step is the shared M-step above.

\section{Experiment Details and Additional Results}
\label[appendix]{app:experiments}

\subsection{Clustering with FYEM}
\label[appendix]{ssec:clustering}

\paragraph{Experiment Details. } 
We draw $1000$ observations from an equally weighted
four-component bivariate GMM. The component means are
$[-1,-1]^\top$, $[0,0]^\top$, $[1,1]^\top$, and
$[1,-1]^\top$, and the corresponding covariance matrices are
$0.11\,\vI$, $0.5\,\vI$, $0.7\,\vI$, and $0.9\,\vI$.
We then draw $100$ additional noise points uniformly from
$[-3,3]^2$; $1100$ observations in total.
We compare the three EM variants described ---standard EM, hard EM, and sparse EM--- on unsupervised clustering of this dataset. The number of components in each model is set to $4$ and the initial parameters (means and covariances) are randomly initialized from a uniform distribution in $[0,\, 0.1]$. Each algorithm runs $200$ iterations, which we verified is more than enough to satisfy a tight convergence criterion.

\begin{figure}[h] 
    \centering
    \includegraphics[width=\linewidth]{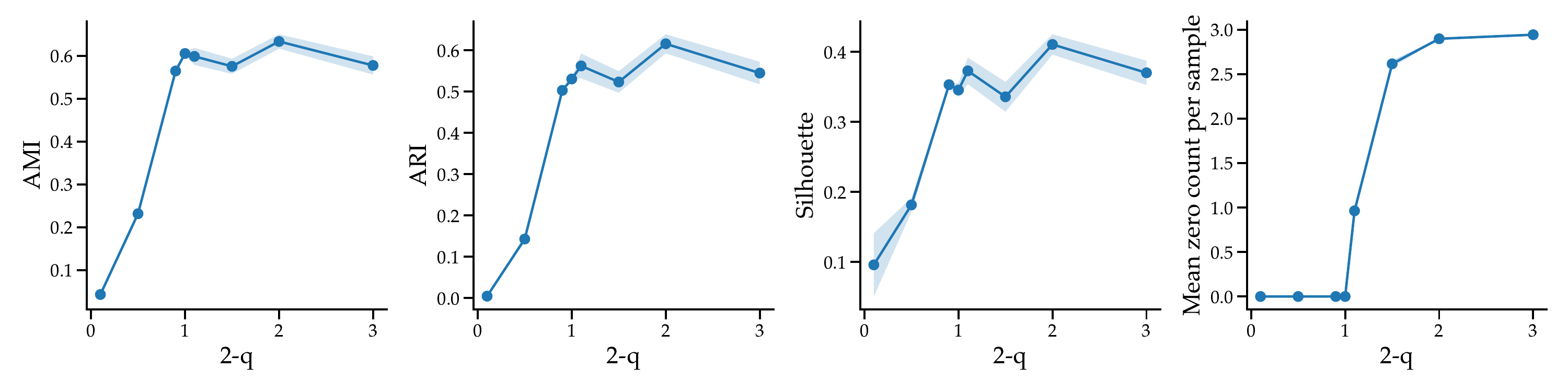}
    \caption{
    We vary the Tsallis entropic index $(2-q) \in \{ 0.1, 0.5, 0.9, 1.0, 1.1, 1.5, 2.0, 3.0\}$ and plot, from left to right the resulting adjusted mutual information, adjusted Rand index, and silhouette score. 
In the \textbf{right-most} panel, we quantify E-step sparsity by counting, for each observation, the number of components assigned zero posterior probability at the final iteration and then averaging across observations. Since each assignment vector sums to one and there are four components, the maximum zero count is three. The shaded regions represent standard errors over five seeds.}
    \label{fig:alpha}
\end{figure}

We evaluate the quality of unsupervised clustering of each algorithm using three metrics: adjusted mutual information \citep{Vinh}, adjusted Rand index \citep{Rand}, and silhouette score \citep{Rousseeuw}. 
Each model is trained and evaluated across five random seeds, and the mean and standard error of each metric are reported in Table~\ref{tab:synthetic-clustering}. In Figure~\ref{fig:alpha}, we vary
\[
2-q
\in
\{0.1,0.5,0.9,1.0,1.1,1.5,2.0,3.0\}
\]
and evaluate the same three metrics. 
We quantify E-step sparsity by the average number of components
assigned zero posterior mass at the final iteration $R$:
\[
\frac{1}{N}\sum_{i=1}^{N}
\#\left\{
k\in\{1,\ldots,K\}:
\pi_i^{(R)}(k)=0
\right\}.
\]


\subsection{MNIST and FMNIST FY VAEs}
\label[appendix]{ssec:vaes}

The objective maximized by standard amortized variational inference \citep{kingma2013auto} w.r.t. $\vpsi$ (encoder parameters) and $\vphi$ (decoder parameters), for a set of $N$, is
\begin{equation}
\sum\limits_{i=1}^N \underset{\text{\vtheta}_i\sim \pi_i}{\mathbb{E}}
\left[ \log p_{\text{\vphi}}(\vx_i\mid\vtheta_i) \right] - \beta \dkl{\pi_i}{\pi_0}
\label{eq:enc_dec}
\end{equation}
where
\[
\pi_i = \pi_{\text{\vpsi}}(\cdot\mid x_i)
= \mathcal{N} \left( \vmu_{\text{\vpsi}}(x_i), \mbox{diag} \left( \vsigma_{\text{\vpsi}}^2(x_i) \right) \right),
\qquad
\pi_0 = \mathcal{N}(\vzero,\vI),
\]
and
\[
p_{\text{\vphi}}(x_i\mid\vtheta_i)
\sim \textnormal{Bernoulli}
\left( \sigma \left( f_{\text{\vphi}}(\vtheta_i) \right) \right).
\]
Here, $\sigma$ denotes the sigmoid function, and $f_{\text{\vphi}}$ is a neurally parametrized decoder function.

\paragraph{Experiment Details.} 
We train VAEs with two-layer (512, 256) neural networks for both encoder and decoder, with a symmetric design. All models whose results are reported in Table~\ref{tab:beta-Gaussian-vae} are trained for 50 epochs using a batch size of 64 and the Adam optimizer with a learning rate of $5 \times 10^{-5}$. The entropic regularizer is weighted by a factor of $\beta = 0.01$ which prevents posterior collapse. We used \cite{vae_torch} as our baseline, keeping the grayscale pixel intensities in $[0,1]$, and finetuned the learning rate and $\beta$ using Optuna \citep{optuna2019}, keeping resources among compared models equal. In all cases $\beta=0.01$ ensures that the regularization term is nontrivially small. In all cases models converge based on training objective curves, and we empirically verified that for $q$ and $q'$ close to 1, training and validation behavior was similar to the standard VAE. We omit these plots because it is misleading to plot objectives that are not numerically comparable to each other. It would also be misleading to compare the expected loss term with the regularization term, as these are also on different scales, e.g. expected log likelihood compared to $\Omega_2(\pi) = 1/2 ||\pi||_2^2 -1$; compare to the log scale of both terms of \cref{eq:enc_dec}. 

We report the $l_1$ reconstruction error in Table~\ref{tab:beta-Gaussian-vae}, i.e. the absolute distance between the model's reconstructed samples and the original input images, evaluated on the test split for both MNIST and FashionMNIST datasets. 
We evaluate $l_1$ reconstruction error between the grayscale input and the sigmoid or binary entmax reconstruction, using one latent sample per test image. 
Absolute errors are summed over pixels and averaged over test images. 
Neither inputs nor reconstructed pixel intensities are binarized.
The reconstruction error is a standard and interpretable metric that is comparable across the VAE alternatives we consider. The FYELBO is log scale only in the standard VAE case. 
In the non-Shannon models we consider, the latent and observation generators have entropic indices $2-q>1$ and $2-q'>1$, equivalently deformation indices $q<1$ and $q'<1$. 
The first row of Table~\ref{tab:beta-Gaussian-vae} corresponds to the standard VAE with a Gaussian latent variable. The Biweight and Epanechnikov densities are bounded support $q$-Gaussian latent distributions with $q=\frac12$ and $q=0$, induced by $\Omega_{3/2}$ and $\Omega_2$, respectively. The best reconstruction errors are obtained with latent $q$-Gaussian posteriors combined with $2$-entmax observation models, corresponding to $q'=0$.

\end{document}